\documentclass[fleqn,10pt]{wlpeerj}
\usepackage[caption=false,farskip=0pt,labelfont={bf}]{subfig}
\usepackage{diagbox}
\title{Dynamic Resolution Guidance for Facial Expression Recognition}
\author[1]{Songpan Wang}
\author[1]{Xu Li}
\author[1]{Tianxiang Jiang}
\author[1]{Yuanlun Xie}
\affil[1]{School of Computer Science and Engineering, University of Electronic Science and Technology of China, Chengdu 611731, China}
\corrauthor[1]{Songpan Wang}{oujieww6@gmail.com}

\begin{abstract}
    
Facial expression recognition (FER) is vital for human-computer interaction and emotion analysis, yet recognizing expressions in low-resolution images remains challenging. This paper introduces a practical method called Dynamic Resolution Guidance for Facial Expression Recognition (DRGFER) to effectively recognize facial expressions in images with varying resolutions without compromising FER model accuracy. Our framework comprises two main components: the Resolution Recognition Network (RRN) and the Multi-Resolution Adaptation Facial Expression Recognition Network (MRAFER). The RRN determines image resolution, outputs a binary vector, and the MRAFER assigns images to suitable facial expression recognition networks based on resolution. We evaluated DRGFER on widely-used datasets RAFDB and FERPlus, demonstrating that our method retains optimal model performance at each resolution and outperforms alternative resolution approaches. The proposed framework exhibits robustness against resolution variations and facial expressions, offering a promising solution for real-world applications.

\end{abstract}

\begin{document}

\flushbottom
\maketitle
\thispagestyle{empty}

\section{Introduction}

Facial expression recognition (FER) is an essential task in video analysis and image understanding, with widespread applications in various fields \cite{tang2019design,lukas2016student,hilles2017knowledge}.

Traditional research on FER has mostly focused on single-subject images, with limited attention given to crowd-scene images. In recent years, FER methods have evolved by employing Convolutional Neural Network (CNN)-based backbone networks for robust feature extraction. Facial expression classification is usually achieved using fully connected layers, Support Vector Machines (SVM), and other similar approaches. Notably, networks such as ResNet~\cite{he2016deep}, Inception network~\cite{szegedy2015going}, and others have demonstrated impressive feature extraction capabilities, leading to satisfactory results when training models with individual and static facial images as input.

However, real-world crowd scenes present numerous challenges for FER. One primary challenge is the prevalence of low-resolution images, which can cause a loss of vital feature information when extracted using traditional methods, leading to decreased discrimination capabilities. Additionally, as the resolution declines, the feature distribution shifts, posing another hurdle for FER in crowd scenes. Critically, in crowd scenes, facial images of different individuals vary in size (as shown in Fig. 1), presenting a significant challenge in achieving high performance with a single FER model. The reduction in image resolution can be traced back to limitations in camera equipment quality and the distance between the subject and the lens. As a result, captured facial images display varying sizes (keys). Fig. 2 illustrates that when the down-sample scale is smaller than x7, the disparity in features between low-resolution and high-resolution versions of the same image becomes more pronounced than the average distance between dissimilar images.

Image super-resolution (ISR) technology can recover high-resolution images with abundant details from low-resolution images, as demonstrated in previous studies~\cite{zhang2018residual,lim2017enhanced,dai2019second,hu2019meta,zhang2018image,lai2017deep}. In some instances, ISR methods have been applied to enhance low-resolution images to improve outcomes in FER tasks~\cite{liu2020facial,cheng2017robust}. However, earlier studies~\cite{jing2020feature} have mainly concentrated on enhancing model accuracy at a fixed resolution, which can restrict a model's adaptability to data with varying resolutions. Nonetheless, there has been a lack of attention devoted to the real-world application of low-resolution facial expression recognition algorithms, few studies have concentrated on the application of expression classification models at varying resolutions.

To optimize the model's performance, guidance is often required to ensure the model can effectively manage data at different resolutions. 
Due to the inherent characteristics of convolutional neural networks, it is challenging to apply them to a wide range of data with different resolutions simultaneously. As a result, using a model trained on a specific resolution or one that has been adapted to incorporate varying resolutions directly may not yield optimal performance. To address the aforementioned challenges, this paper initially investigates adaptation algorithms at varying resolutions and confirms that it is quite difficult to employ a single model for handling facial expression recognition problems across different resolutions. Subsequently, we propose a Dynamic Resolution Guidance(DRG) framework that can automatically identify the resolution of the input facial image and forward it to the corresponding FER network for recognition. To determine the resolution of each face, a Resolution Recognition Network(RRN) is introduced. Finally, we validate our proposed framework on several widely used facial expression datasets. The experimental results demonstrate that our algorithm can achieve superior performance.

\begin{figure}[h]%
\centering
\includegraphics[width=0.9\textwidth]{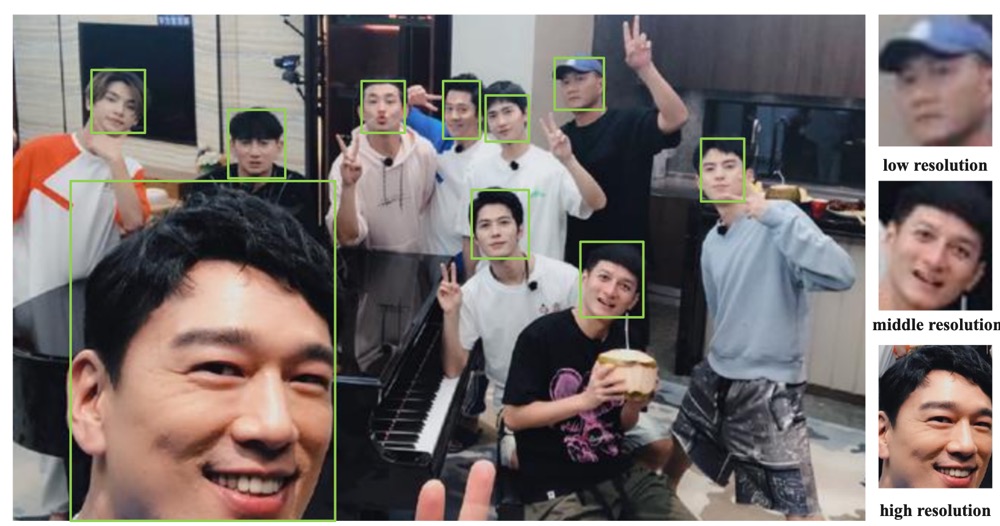}
\caption{This is a group photo featuring Chinese celebrities. Due to the shooting angle and distance, the resolution of each individual's face varies. We have selected facial images of three celebrities on the right for an intuitive visual comparison. The images demonstrate the differences in clarity at three distinct resolutions: high, medium, and low.}\label{fig1.0}
\end{figure}

\section{Related Work}

Conventional convolutional neural network (CNN) approaches~\cite{tumen2017facial} typically consider only standard-sized images and do not employ edge-aware feedback mechanisms. However, practical scenarios, such as surveillance camera footage, frequently involve low-resolution facial images. Often, these real images have lower resolution compared to the experimental dataset. One major obstacle in recognizing facial expressions in low-resolution images is that current facial expression recognition networks primarily focus on images of ideal size, resulting in a substantial decrease in recognition accuracy as image resolution decreases.  
Jie Shao et al.~\cite{shao2021fcnn} introduced an edge-aware feedback convolutional neural network (E-FCNN) for recognizing facial expressions in tiny low-resolution images. The E-FCNN incorporates feedback connections between convolutional layers and employs edge-aware convolutional layers to capture detailed information within the images. Experimental evaluations were conducted using downsampled images from four facial expression datasets (CK+, FER2013, BU-3DFE, RAF-DB), revealing superior performance and higher accuracy in the corresponding downsampling magnifications. However, the reconstructed images through super-resolution (SR) techniques still exhibited blurriness.

In the context of single image super-resolution (SISR) frameworks, Wu Gang et al.~\cite{wu2021practical} investigated sample construction and feature embedding, proposing a task-friendly embedding network based on adversarial learning. This network facilitated better reconstruction of lost high-frequency information by generating information-rich positive samples and challenging negative samples in the frequency space. This approach enhanced the model's adaptability to low-level tasks requiring rich texture and contextual information, thereby advancing research in single-image super-resolution (SISR). However, the single-image super-resolution model struggles to handle multi-scale images effectively, and the reconstruction performance is significantly influenced by the reduced resolution of the input image. Consequently, it fails to guarantee the discriminative sufficiency of recovered features for specific tasks such as object detection and expression classification.

Nan Fang et al.~\cite{nan2022feature} proposed a feature super-resolution-based FER method. Their approach employed a novel GAN training strategy that directed the model's attention toward samples that were difficult to classify into the corresponding categories. Experimental evaluations were conducted using high-resolution images of RAF-DB with varying downsampling factors (ranging from x2 to x8). The results demonstrated that the loss reweighting strategy effectively accelerated the training process while keeping other parameters fixed. However, the experiment did not provide specific data on the experimental results of low-resolution facial image classification.

The recognition of facial expressions in multi-scale and low-resolution images has long been overlooked due to the absence of crowd scene images in public datasets. Previous researchers predominantly treated multi-scale recognition as a separate task, necessitating the training of a distinct model for each scale. This approach proved inefficient, particularly at low resolutions.
To overcome these limitations, we propose a Dynamic Resolution Guidance(DRG) framework in this study. This framework incorporates a Resolution Recognition Network(RRN) that predicts the resolution of an image in comparison to previous works and subsequently feeds it into the corresponding resolution network. 
\section{Single Model Adaptation}
We explored various methods to enable a single FER network model to effectively analyze multi low-resolution facial expression images. 

\textbf{Multi Scale Training (MSTrain).} This approach represents a straightforward and essential methodology~\cite{ou2020efficient}. By incorporating data augmentation techniques into the training process, a diverse range of low-resolution facial expression image data can be effectively simulated. The underlying objective is to enable the neural network to effectively adapt to these varying resolutions, thereby facilitating targeted training specifically tailored to a particular resolution setting.
Regrettably, despite its initial promise, this method did not yield the desired outcome. In fact, it resulted in a noticeable decrease in the model's accuracy across different resolutions.

\textbf{Domain Adaptive.}
Domain adaptation~\cite{zhang2019bridging} has emerged as a prominent research area in recent years. It primarily addresses the effects of data distribution discrepancies on the performance of machine learning models. This concept can be applied to the challenge of multi-scale, low-resolution facial expression recognition. Although data resolutions may vary, leading to distribution biases, the representations employed for classification exhibit similarities.
Domain adaptation primarily employs feature vectors to accomplish two distinct recognition tasks: the original task of expression recognition and domain recognition, where different resolutions represent separate domains. The training process is based on adversarial learning, with the objective of enabling the feature encoder to deceive the domain recognizer. This makes it challenging for the domain recognizer to differentiate between the source domain (presumably using 1x magnification images) and the target domain (presumably using images downsampled by a factor of 2).
Upon testing, we discovered that this approach cannot prevent a decline in accuracy.

\textbf{Resolution-aware BN.}
Zhu et al.~\cite{zhu2021dynamic} conducted research and determined that varying resolution data exhibit distribution shifts, making it challenging for a single neural network model to adapt to multiple resolutions concurrently. To address this issue, they introduced multiple independent BatchNorm modules in parallel following the convolutional layer, as opposed to the traditional approach of using a single convolutional layer and BatchNorm module. For images of differing resolutions, each respective BatchNorm layer performs an independent normalization operation on the current data. This technique enables the authors to project data from various resolutions into a consistent latent space, thereby reducing the distribution discrepancy between different resolutions.
We applied this method to facial expression recognition as well. However, our experimentation showed that this approach does not alleviate the issue of a single model's recognition accuracy degradation when applied to multiple-resolution scenarios.

Upon investigating several aforementioned techniques, we discovered that certain methods, while theoretically promising, do not yield satisfactory results in practical applications.
\section{Methodology}
\begin{figure}[htbp]%
\centering
\includegraphics[width=0.9\textwidth]{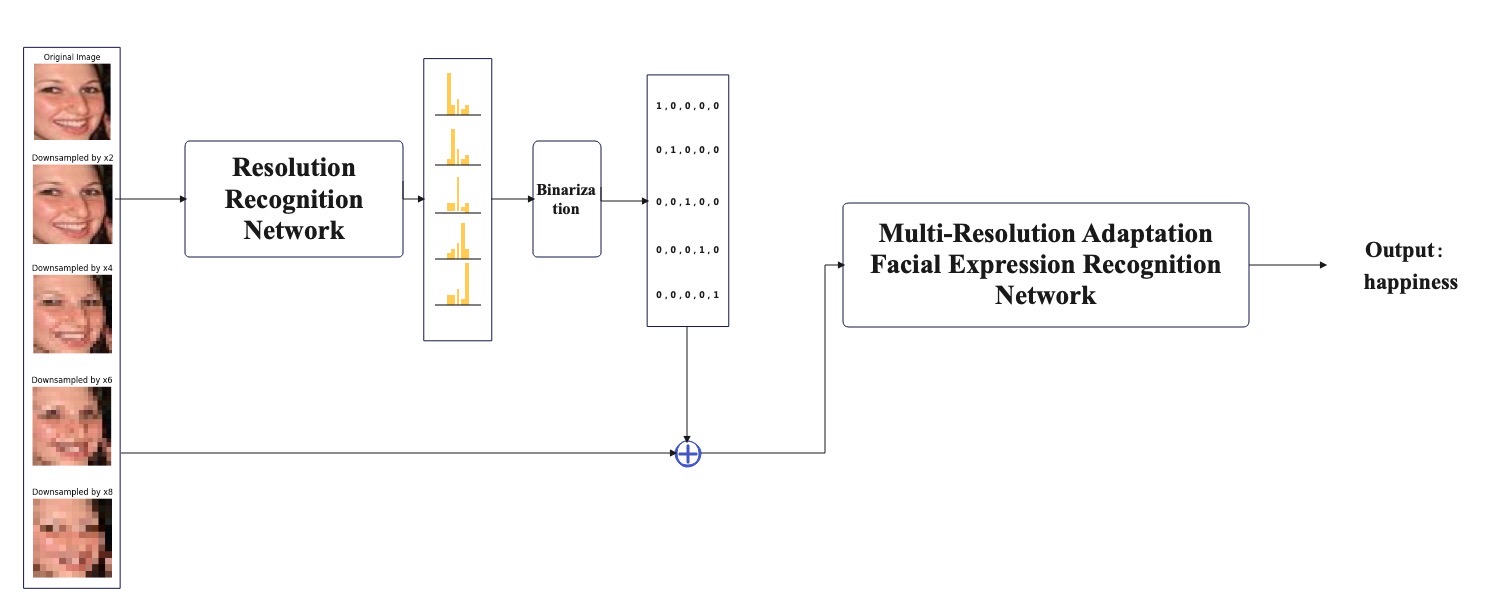}
\caption{This is pipeline of our proposed method.}\label{fig1}
\end{figure}

Our proposed method in this paper is both simple and practical, Dynamic Resolution Guidance for Facial Expression Recognition (DRGFER). By utilizing our framework, it is possible to achieve end-to-end automatic recognition of various low-resolution facial expressions without compromising the accuracy of the FER model.
As depicted in Figure \ref{fig1}, our proposed framework comprises two stages. Initially, 
the Resolution Recognition Network(RRN) is utilized to determine the resolution of the input facial expression image. Subsequently, a binarization operation is performed to convert the network's output into a 0-1 vector. The original image, denoted as $I_i$, and the binarized vector, denoted as $I_i^b$, are then fed into the Multi-Resolution Adaptation Facial Expression Recognition Network(MRAFER) as a pair. The network model will automatically select the appropriate facial expression recognition network according to $I_i^b$ and ultimately output the recognition result.

\subsection{Resolution Recognition Network}

\begin{figure}[htbp]%
\centering
\includegraphics[width=0.9\textwidth]{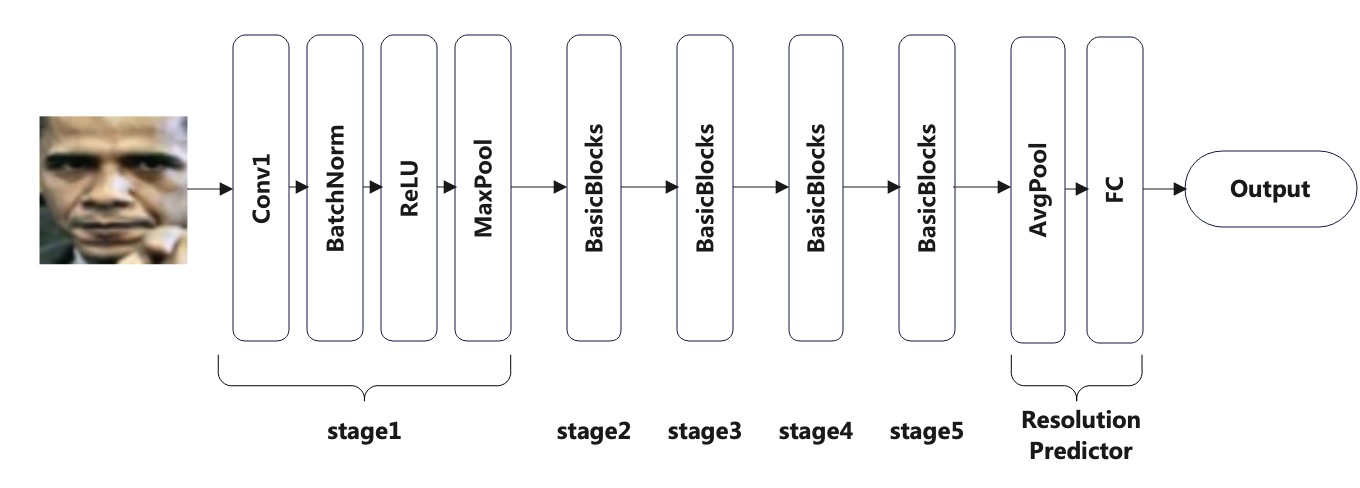}
\caption{Our RRN is based on Deep residual network(ResNet18)}
\label{fig2}
\end{figure}

To address the problem of facial expression recognition at different resolutions, we first propose a resolution recognition network to guide the subsequent recognition model for more accurate classification of facial expressions.

\begin{figure}[h]%
\centering
\includegraphics[width=0.5\textwidth]{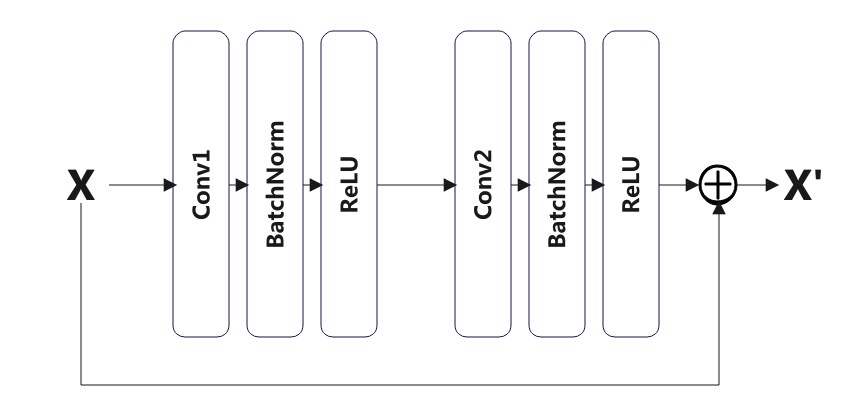}
\caption{BasicBlock}
\label{fig3}
\end{figure}

We employ the ResNet18, which use the same architecture as the FER network used later in the process. The structure of the network, as shown in Fig.\ref{fig2}, is divided into six parts: stage 1, stages 2-5, and the final resolution predictor. The first stage consists of a convolutional layer, batch normalization layer, ReLU, and max-pooling layer, which extract low-level features from the image while performing downsampling twice. As a result, after the first stage computation, the feature map resolution is only 1/4 of the input image. The subsequent four stages are composed of BasicBlocks, with two BasicBlocks in each stage. Each BasicBlock is a residual module, as illustrated in Fig.~\ref{fig3}, 
consisting of two convolution operations, after each convolution operation, BatchNorm is employed to normalize the data, followed by the ReLU operation, which performs nonlinear mapping on the data. The most crucial aspect is that the input and output of the module are added together, utilizing the residual concept to guide the weights in the module during training.
Among stages 2-5, only stage 2 does not include any downsampling operations, while the others do. Finally, the resolution predictor consists of an average pooling layer and a fully connected layer. The feature map is converted into a feature vector through average pooling, followed by a prediction to determine the resolution of each image.

The model's output consists of an unnormalized vector for classification purposes,  which is a kind of probability distribution of the MLP, we can donated it as $\mathbf{\pi^i}$ for $i^{th}$ image.

\textbf{Loss.} Essentially, our RRN is a classification task. Therefore, we employ Softmax to normalize the output vector and utilize the cross-entropy loss function to guide the learning process for this specific component, by eq.\ref{eq1} and eq.\ref{eq2}.

\begin{equation}
\label{eq1}
    \hat{\textbf{y}^i_j} = \frac{\exp(\pi_j^i)}{\sum_{k=1}^{C} \exp(\pi_k^i)}
\end{equation}

\begin{equation}
\label{eq2}
    L_{RRN}(\textbf{y}^i, \hat{\textbf{y}}^i) = -\sum_{j=1}^{C} y^i_j \log(\hat{y}^i_j)
\end{equation}

\textbf{Binarization.} The binarization operation is employed to convert the vector output by the network, as the vector output by the RRN cannot be directly used by our MRAFER. This binarization operation does not rely on a preset threshold, instead, it is based on the maximum value. In this approach, the element with the maximum value in the vector is set to 1, while all other elements are set to 0, the whole process can be defined as the following formula:
\begin{equation}
I^b_{i,j} =
\begin{cases}
    1 & \text{if } I^b_{i,j} = \mathbf{max}(I^b_i) \\
    0 & \text{otherwise}
\end{cases}
\end{equation}
In this equation, $I^b_{i,j}$ represents the resolution $j$ for image $I_i$, and superscripts represent binarization. The operation sets the element $I^b_{i,j}$ to 1 if it is equal to the maximum value in the corresponding  $I^b_i$, and to 0 otherwise. 
This results in a binary vector,  which can be effectively utilized by the MRAFER to select the appropriate FER\_Block for facial expression recognition.

\subsection{Multi-Resolution Adaptation Facial Expression Recognition Network}

\begin{figure}[h]%
\centering
\includegraphics[width=0.9\textwidth]{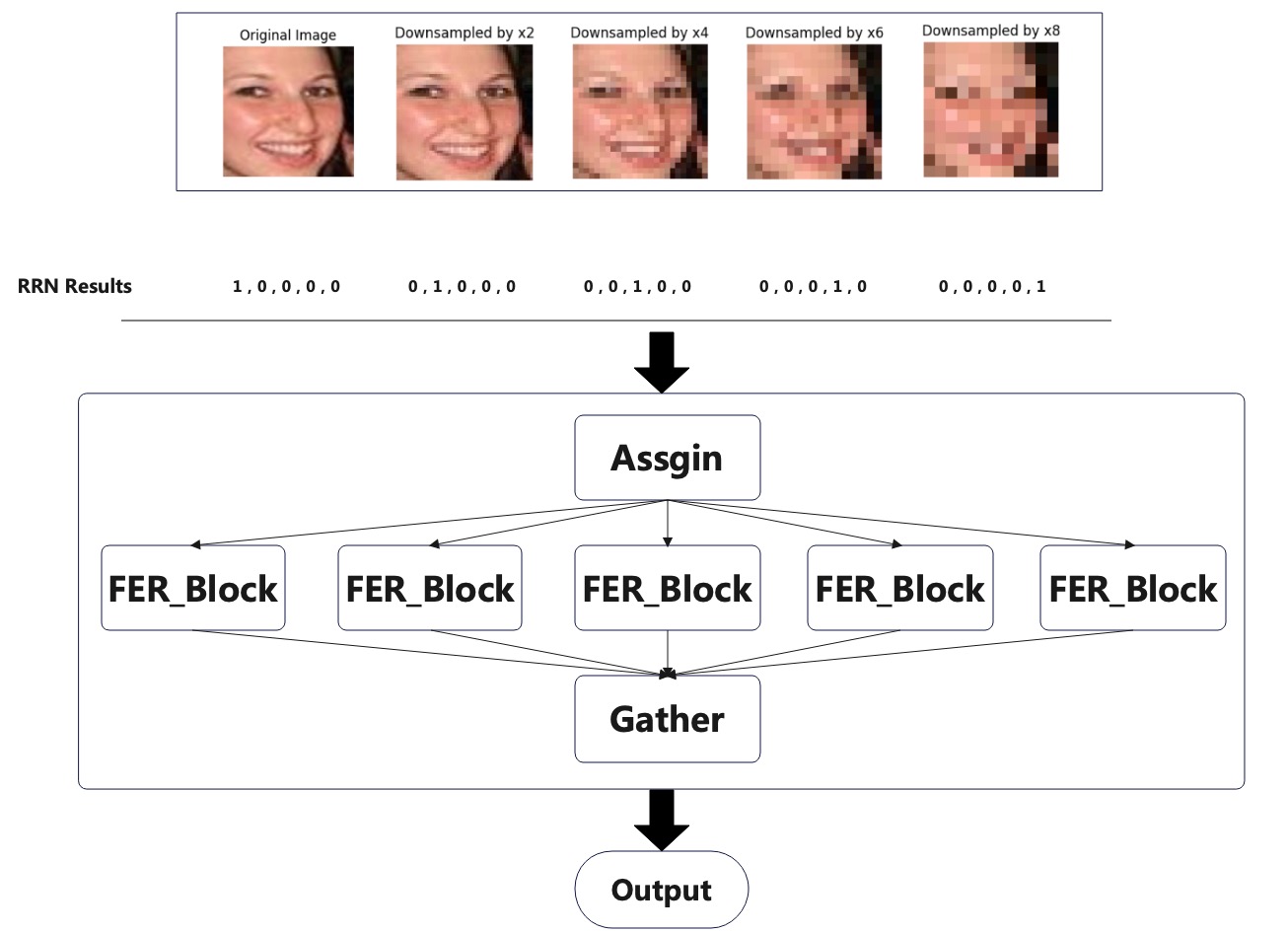}
\caption{Multi-Resolution Adaptation Facial Expression Recognition Network.}
\label{fig4}
\end{figure}
As illustrated in Fig.\ref{fig4}, our Multi-Resolution Adaptation Facial Expression Recognition(MRAFER) is comprised of three main components: Assign, FER\_Block, and Gather. 
First, our Assign module traverses the resolution predictions in the entire batch data, then combines images with different resolutions into new batch data, and sends different batches to the corresponding FER. The network structure of our FER module is shown in Figure 3. The structure is the same as that of RRN. The only difference is that the number of outputs of the last fully connected layer is different. This is related to the dataset used. 
Finally, we need to splice the batch data predicted by each FER module and use the Gather operation.

\begin{itemize}
    \item [1) ] Traverse the resolution predictions $I_i^b$, in the entire batch data, $B$.
    \item [2) ] Grouping images with different resolutions into a new batch data,  $B'={B_1, B_2, ..., B_k}$, $k$ is the number of FER\_Blocks. 
    \item [3) ] Send $B'$ to the corresponding FER\_Blocks.
    \item [4) ] Process the images through the FER\_Block (network structure shown in Fig.\ref{fig3}.
    \item [5) ] Obtain the predictions $B_k^p$ for each $B_k$ and Splice the batch data predicted by each FER\_Block into a single output $B^p$. 
\end{itemize}

By the above steps, our MRAFER can effectively handle and process facial expression images with varying resolutions. The FER\_Blocks, tailored to different resolutions, can provide more accurate predictions, and the Gather operation ensures that the final output is combined correctly for further analysis or processing.
\begin{gather}
\label{4-7}
    B' = \mathbf{Assgin}(B) \\
    B_k = \{I_i|I_i^b[k]=1\} \\
    B^p = \mathbf{Gather}(\{B_1^p, B_2^p, ..., B_k^p\}) \\
    B^p[I_i^b[k]] = B_k^p
\end{gather}
We can use eq.4\-7 to define two operations, Assign and Gather. eq.5 provides the details of eq.4, $B_k$ represents the set of images $I_i$ with their binarized vector $I_i^b[k]$ equal to 1. This indicates that they belong to the $k$-th resolution group. And eq.7 provides the details of eq.6,  $B^p[I_i^b[k]]$ denotes the prediction for each image $I_i$ in the $k$-th resolution group. The Gather operation assigns the prediction from the corresponding resolution group $B_k^p$ to the final output $B^p$.

\section{Experiment}
\subsection{Dataset}

To assess the performance of expression recognition, we utilize the RAF-DB~\cite{li2017reliable} and FERPlus~\cite{barsoum2016training} datasets in our experiments. \textbf{RAF-DB} was compiled using various search engines, and approximately 40 annotators independently labeled each image. The dataset comprises 15,339 images labeled with seven basic emotion categories, with 12,271 designated for training and 3,068 for validation. \textbf{FERPlus} is an extension of FER2013, as used in the ICML 2013 Challenges. It is a large-scale dataset collected via the Google search engine, containing 28,709 training images, 3,589 validation images, and 3,589 test images, each resized to $48\times 48$ pixels. The dataset includes an additional class, contempt, resulting in a total of 8 classes. The overall sample accuracy serves as the performance metric.

\begin{figure}[h]%
\centering
\includegraphics[width=0.9\textwidth]{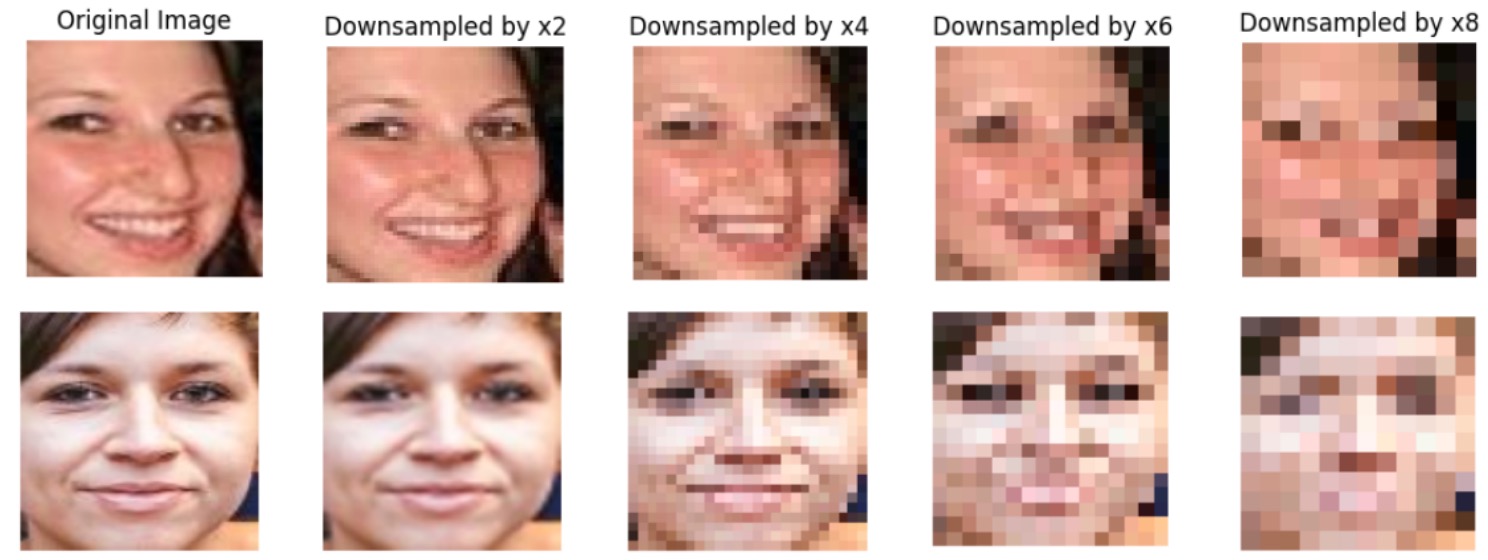}
\caption{Two images from the RAF-DB dataset are visualized. The first column presents the original size image, while the subsequent columns display the magnified images after reducing the corresponding magnification.}
\label{fig5}
\end{figure}

Similar to most super-resolution studies, we apply a bicubic kernel function to downsample high-resolution images and obtain low-resolution counterparts. The original input size is 100 × 100 pixels, and we achieve low-resolution images by employing integer down-sample factors of x2, x4, x6, and x8. Consequently, the total pixel count is reduced to 1/4, 1/16, 1/36, and 1/64. Fig. \ref{fig5} displays several examples post-downsampling.

\subsection{Experiment Setup}

To evaluate the performance of DRGFER, we compare it with simple max-pooling, average-pooling strategies, and the previously mentioned multi-scale augmentation training, domain adaptation, and resolution-aware batch normalization methods. We assess the results based on total accuracy. The input size is set to $100 \times 100$. For data augmentation, we employ minimal settings, including resizing to 224, random horizontal flip, and normalization. We implement the code using PyTorch. All experiments have a batch size of 256, a learning rate of 3e-4, and use the Adam optimizer for 80 iterations. Our experiments utilize an Nvidia 1080Ti GPU.

\subsection{Facial Expression Recognition Result}

\begin{table}[ht]
\centering
\caption{Results of total accuracy on RAF-DB with different resolution. Best results are \textbf{highlighted}.}
\begin{tabular}{|c|c|c|c|c|c|c|}

\hline
\diagbox{Ratio}{Accuracy}{Methods} & Mean  & Max & RA-BN& DA & MSTrain &DRGFER\\
\hline
x1 & 86.11\% & 86.64\% & 86.96\%& 81.10\% &85.88\%&\textbf{89.24}\%\\
x2 & 86.34\% & 86.28\% & 86.73\%& 81.10\% &85.91\%&\textbf{88.23}\%\\
x4 & 83.93\% & 82.92\% & 84.41\%& 77.22\% &84.35\%&\textbf{85.30}\%\\
x6 & 75.98\% & 75.62\% & 80.18\%& 69.04\% &80.93\%&\textbf{81.91}\%\\
x8 & 66.85\% & 70.47\% & 76.43\%& 61.01\% &77.18\%&\textbf{77.35}\%\\
\hline
\hline
Mean & 79.84\% & 80.38\% &82.94\%& 73.89\% & 82.85 \%&\textbf{84.41}\%\\
\hline

\end{tabular}
\label{t1}
\end{table}
The Table.\ref{t1} presents the total accuracy results of six different face emotion recognition methods (Mean, Max, RA-BN, DA, MSTrain, and DRGFER) on the RAF-DB dataset, with varying input image resolutions represented by down-sampling ratios (x1, x2, x4, x6, and x8). The best results in each row are highlighted in bold.


For x1 down-sampling ratio, DRGFER achieves an accuracy that surpasses the 89.24\% mark, trailed by RA-BN with slightly over 86.96\%, and MSTrain at 85.88\%. Mean and Max techniques yield comparable results with an accuracy just above 86.11\% and approximately 86.64\%, respectively, while DA records the lowest accuracy at just 81.10\%.


At the x2 ratio, DRGFER maintains its leading position with an accuracy just over 88\%, registering at 88.23\%. The RA-BN method accomplishes 86.73\% accuracy, whereas MSTrain exhibits a slight dip to 85.91\%. Mean and Max methods portray nearly identical performance, yielding 86.34\% and 86.28\% accuracy, respectively. DA persists at the bottom with 81.10\%.


When the down-sampling ratio increases to x4, DRGFER continues its dominant position with an accuracy of 85.30\%. MSTrain and RA-BN perform comparably with accuracies of 84.35\% and 84.41\%, respectively. The Mean and Max methods experience a drop in their performance, registering 83.93\% and 82.92\% accuracy, respectively, while DA further slips to 77.22\%.


At the x6 ratio, DRGFER leads with a top accuracy of 81.91\%, followed by MSTrain at 80.93\%. RA-BN exhibits a lower accuracy of 80.18\%, trailed by the Max method at 75.62\% and the Mean method at 75.98\%. The DA method shows a substantial plunge in performance with an accuracy of 69.04\%.

For the highest down-sampling ratio of x8, DRGFER once again achieves the best accuracy at 77.35\%. MSTrain and RA-BN follow with 77.18\% and 76.43\% accuracy, respectively. The Max method has an accuracy of 70.47\%, and the Mean method drops to 66.85\%. The DA method's performance declines further to 61.01\%.

In the last row, the table displays the mean accuracy for each method across all down-sampling ratios. DRGFER achieves the highest mean accuracy of 84.41\%, followed by MSTrain at 82.85\%, RA\-BN at 82.94\%, Max at 80.38\%, Mean at 79.84\%, and DA at 73.89\%. The results indicate that the DRGFER method consistently outperforms the other tested approaches across various input image resolutions.

From Table.\ref{t1}, it can be observed that other methods suffer from a significant loss of accuracy when the downsampling factor is relatively small. For RA-BN and MSTrain, training with a mixture of different low-resolution data leads to a certain improvement in the recognition performance of these models under lower-resolution scenarios. Our proposed DRGFER effectively preserves the best accuracy results at each downsampling factor, ensuring optimal facial expression recognition accuracy across different scales.

\begin{table}[ht]
\centering
\caption{Results of total accuracy on FERPlus with different resolutions. The best results are \textbf{highlighted}.}
\begin{tabular}{|c|c|c|c|c|c|c|}
\hline
\diagbox{Ratio}{Accuracy}{Methods} & Mean  & Max & RA-BN& DA & MSTrain &DRGFER\\
\hline
x1 & 82.66\% & 80.86\% & 83.85\%& 80.77\% &82.42\%&\textbf{84.12}\%\\
x2 & 83.18\% & 81.24\% & 83.27\%& 80.37\% &82.54\%&\textbf{83.76}\%\\
x4 & 80.05\% & 79.53\% & 82.02\%& 71.90\% &81.47\%&\textbf{82.23}\%\\
x6 & 70.94\% & 73.41\% & 78.48\%& 49.26\% &77.50\%&\textbf{78.48}\%\\
x8 & 62.01\% & 68.04\% & 74.05\%& 40.91\% &75.01\%&\textbf{75.01}\%\\
\hline
\hline
Mean & 75.77\% & 76.61\% &80.33\%& 64.64\% & 79.79 \%&\textbf{80.72}\%\\
\hline
\end{tabular}
\label{t2}
\end{table}
Table \ref{t2} presents the total accuracy results of six different facial emotion recognition methods (Mean, Max, RA-BN, DA, MSTrain, and DRGFER) on the FERPlus dataset, with varying input image resolutions represented by down-sampling ratios (x1, x2, x4, x6, and x8). The best results in each row are highlighted in bold.


For the x1 down-sampling ratio,  DRGFER outperforms, achieving the highest accuracy just above 84\%, exactly at 84.12\%, followed by RA-BN with an accuracy close to 84\%, settling at 83.85\%, and MSTrain, exceeding 82\%, with 82.42\%. Mean and Max methods yield accuracies of 82.66\% and 80.86\%, respectively, while DA attains 80.77\%.


At the x2 ratio, DRGFER sustains its supremacy, recording an accuracy of 83.76\%. The RA-BN method accomplishes an accuracy of 83.27\%, while MSTrain exhibits a slight surge to 82.54\%. Mean and Max methods display improved performance, offering accuracies of 83.18\% and 81.24\%, respectively. DA, on the other hand, experiences a minor drop to 80.37\%.

When the down-sampling ratio increases to x4, DRGFER still outperforms the other methods with an accuracy of 82.23\%. MSTrain and RA-BN have similar performance at 81.47\% and 82.02\% accuracy, respectively. The Mean and Max methods experience a decline in their accuracy, with 80.05\% and 79.53\% accuracy, respectively, while DA further decreases to 71.90\%.


At the x6 ratio, both DRGFER and RA-BN claim the highest accuracy of 78.48\%. MSTrain trails with an accuracy of 77.50\%, followed by the Max method at 73.41\% and the Mean method at 70.94\%. DA method's performance significantly plummets to 49.26\%.


For the highest down-sampling ratio of x8, both DRGFER and MSTrain take the lead, securing an accuracy of 75.01\%. RA-BN follows with an accuracy of 74.05\%. The Max method delivers an accuracy of 68.04\%, whereas the Mean method descends to 62.01\%. DA's performance deteriorates further to 40.91\%.


In the last row, the table displays the mean accuracy for each method across all down-sampling ratios. DRGFER secures the top mean accuracy of 80.72\%, trailed by MSTrain at 79.79\%, RA-BN at 80.33\%, Max at 76.61\%, Mean at 75.77\%, and DA at 64.64\%. The findings demonstrate that the DRGFER method consistently surpasses the other methods examined across a variety of input image resolutions.

From Table \ref{t2}, it can be observed that other methods suffer from a significant loss of accuracy when the downsampling factor is relatively small. For RA-BN and MSTrain, training with a mixture of different low-resolution data leads to a certain improvement in the recognition performance of these models under lower-resolution scenarios. Our proposed DRGFER effectively preserves the best accuracy results at each downsampling factor, ensuring optimal facial expression recognition accuracy across different scales.
\subsection{ Ablation study}
\begin{table}[h]
\centering
\caption{Result comparison  with different resolution.}
\resizebox{\textwidth}{!}{
\begin{tabular}{|c|c|c|c|c|c|c|c|c|}
\hline
 & x1 & x2 & x4 & x6 & x8 & x12 & x14 & x16 \\ 
 \hline
x1 & \textbf{89.24}\% & 86.17\% & 77.11\% & 65.41\% & 56.22\% & 37.12\% & 33.21\% & 24.51\% \\
 \hline
x2 & 86.96\% & \textbf{88.23}\% & 79.27\% & 66.30\% & 54.50\% & 30.44\% & 25.46\% & 22.00\% \\
 \hline
x4 & 81.42\% & 83.51\% & 85.20\% & 73.89\% & 59.84\% & 31.75\% & 25.68\% & 22.33\% \\
 \hline
x6 & 66.72\% & 70.76\% & 78.39\% & 80.35\% & 69.82\% & 49.93\% & 41.20\% & 23.37\% \\
 \hline
x8 & 54.27\% & 54.60\% & 64.34\% & 73.50\% & 76.86\% & 55.74\% & 50.23\% & 40.45\% \\
 \hline
x1\&x2 & 88.33\% & 88.07\% & 81.55\% & 68.12\% & 56.19\% & 33.41\% & 29.95\% & 22.75\% \\
 \hline
x1\&x2\&x4 & 87.61\% & 87.48\% & \textbf{85.30}\% & 76.37\% & 64.24\% & 44.56\% & 36.51\% & 22.46\% \\
 \hline
x1\&x2\&x4\&x6 & 86.67\% & 86.41\% & 84.94\% & \textbf{81.91}\% & 73.01\% & 53.85\% & 48.11\% & 28.03\% \\
 \hline
x1\&x2\&x4\&x6\&x8 &85.88\% &85.88\% &84.25\% & 80.96\%& 77.21\%&58.34\% &51.59\% &41.85\%\\
\hline
DRGFER & \textbf{89.24}\% &\textbf{88.23}\% &\textbf{85.30}\% &\textbf{81.91}\% &\textbf{77.35}\% & - & -& -\\
\hline
\end{tabular}
}
\label{t3}
\end{table}

Table.\ref{t3} investigates the impact of data augmentation at different resolutions on models trained with various low-resolution data. It demonstrates that when a model is trained using a single resolution, it achieves the best results only at that specific resolution. However, when multiple resolutions are combined for joint training, the accuracy at some lower resolutions can be improved.
For instance, when training solely at the x8 magnification, the model can only achieve an accuracy of 76.86\%, which is 0.65\% lower than the best accuracy obtained through joint training. This suggests that incorporating multiple resolutions during training can lead to better performance in certain lower-resolution scenarios, as the model becomes more robust in handling variations in input image quality. In this case, similar observations can be made for the x4 and x6 magnifications, where joint training leads to accuracy improvements of 0.1\% and 1.56\%, respectively. These results reinforce the idea that combining multiple resolutions during training can enhance the performance of the model in handling lower-resolution input images. By training jointly on a variety of resolutions, the model becomes more adaptable and robust, leading to improved recognition accuracy across different scales.
\begin{figure}[htbp]
	\centering
	\subfloat[Recognition accuracy for x1. ]{\includegraphics[width=.45\columnwidth]{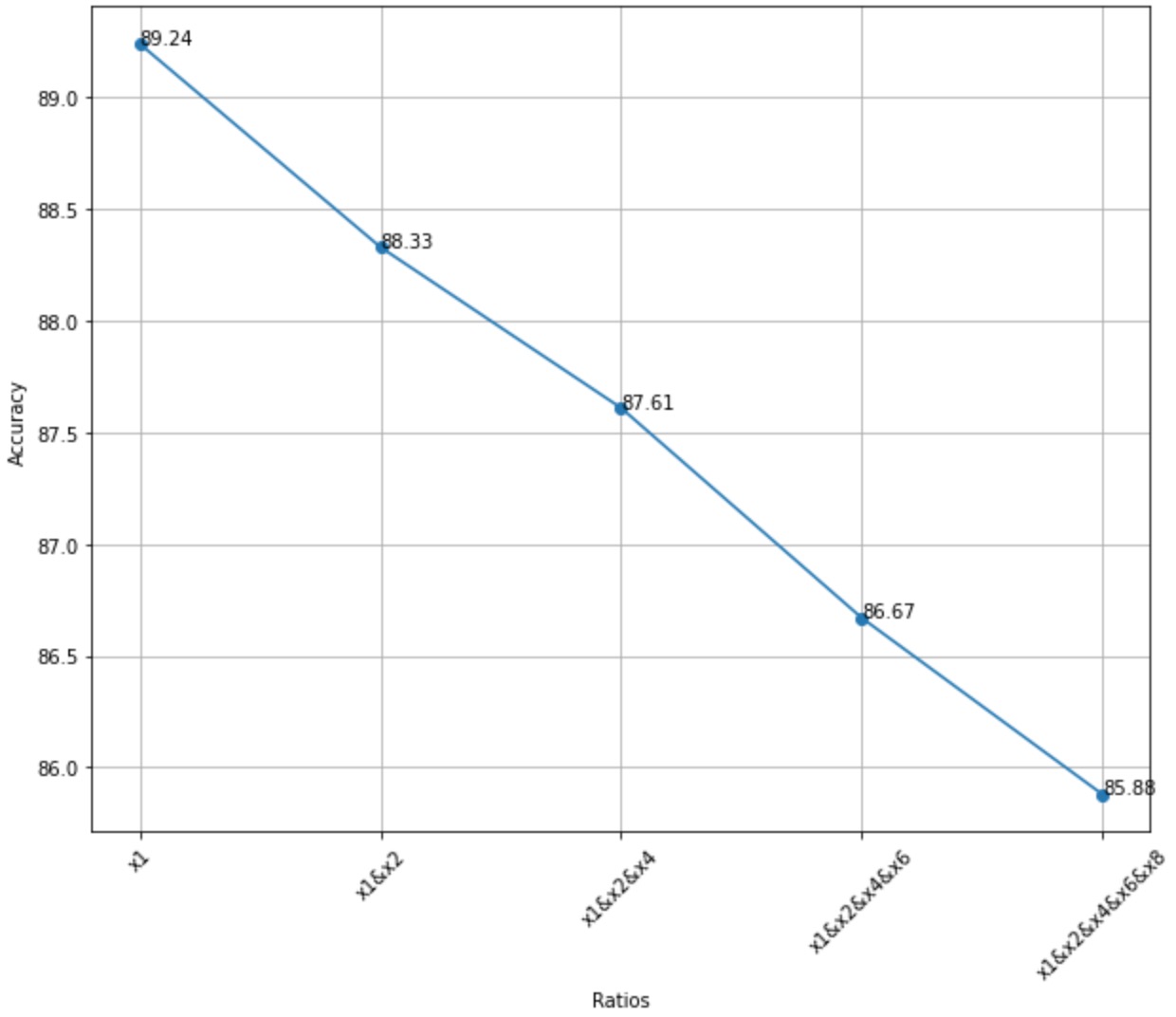}\label{fig21a}}
	\subfloat[Recognition accuracy for x2.]{\includegraphics[width=.45\columnwidth]{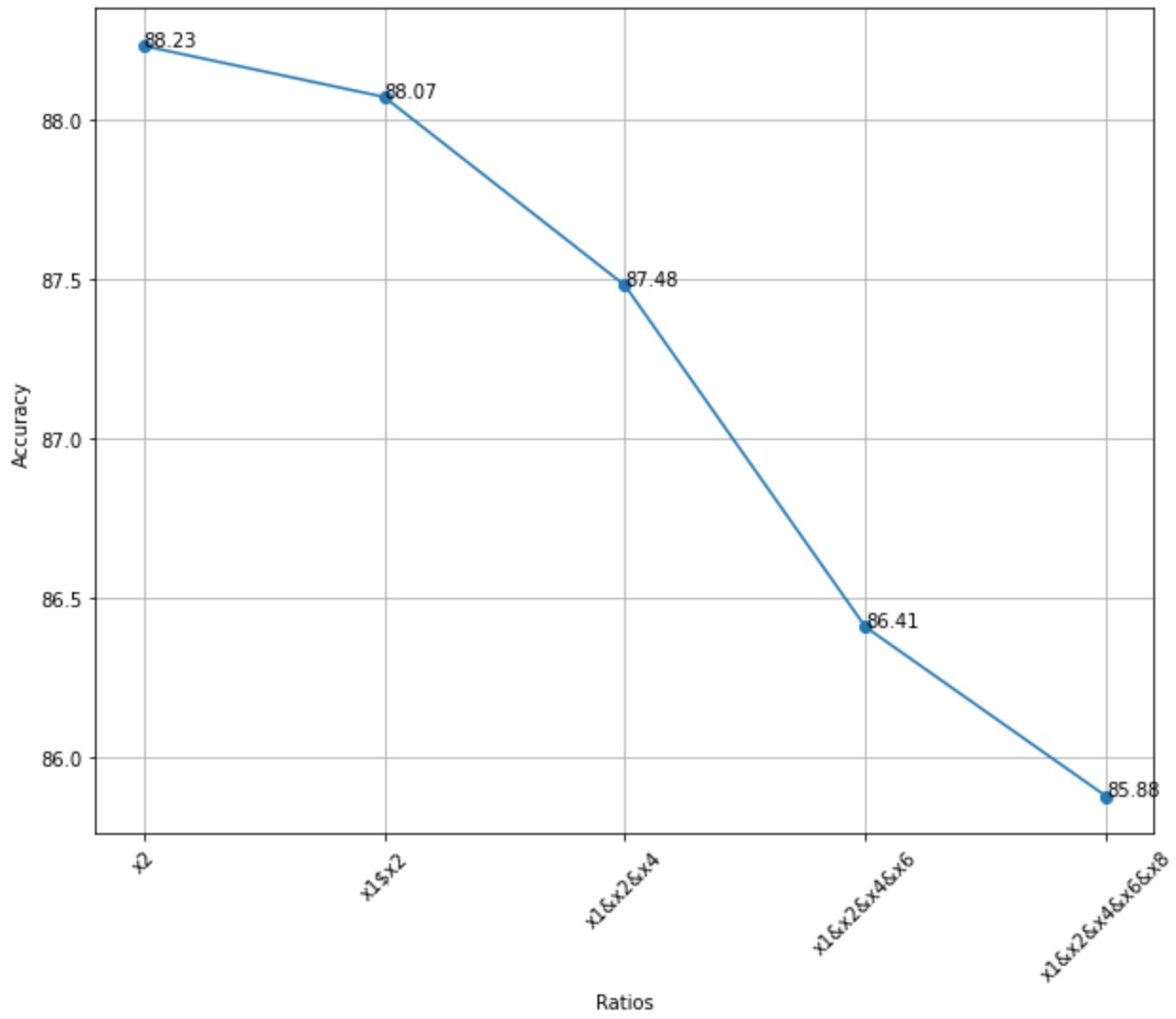}\label{fig21b}} 
 \caption{Comparison of recognition accuracy.}
\end{figure}

Indeed, joint training does not always lead to improvements in relatively high-definition facial expression images. As evidenced by the data at x1 and x2 magnifications, joint training can result in reduced accuracy for both resolutions. Furthermore, the more joint resolutions included in the training, the lower the accuracy becomes, as shown in Fig.\ref{fig21a} and Fig.\ref{fig21b}.
This phenomenon can be attributed to the fact that models trained on low-resolution images may not be as effective in capturing high-resolution features, leading to poorer performance on high-definition input images. Therefore, while joint training can enhance the model's adaptability to lower-resolution images, it may also compromise its performance at higher resolutions. To address this trade-off, it is essential to develop strategies that can strike a balance between the model's accuracy at different resolutions.
\section{Conclusion}
In this paper, we proposed a novel method called Dynamic Resolution Guidance for Facial Expression Recognition (DRGFER) to effectively recognize facial expressions in different low-resolution images without compromising the accuracy of the FER model. Our framework consists of two main components: the Resolution Recognition Network (RRN) and the Multi-Resolution Adaptation Facial Expression Recognition Network (MRAFER).


Our proposed DRGFER framework demonstrates a practical and effective approach to handle and process facial expression images with varying resolutions. The tailored FER\_Blocks for different resolutions ensure accurate predictions, while the Gather operation maintains the integrity of the final output for further analysis or processing. Future work could explore the integration of additional resolution-specific features or network architectures to further improve the performance of the framework across a wider range of resolutions and facial expressions.



\bibliography{mybibliography}

\begin{thebibliography}{}

\bibitem[Barsoum et~al., 2016]{barsoum2016training}
Barsoum, E., Zhang, C., Ferrer, C.~C., and Zhang, Z. (2016).
\newblock Training deep networks for facial expression recognition with crowd-sourced label distribution.
\newblock In {\em Proceedings of the 18th ACM international conference on multimodal interaction}, pages 279--283.

\bibitem[Cheng et~al., 2017]{cheng2017robust}
Cheng, B., Wang, Z., Zhang, Z., Li, Z., Liu, D., Yang, J., Huang, S., and Huang, T.~S. (2017).
\newblock Robust emotion recognition from low quality and low bit rate video: A deep learning approach.
\newblock In {\em 2017 Seventh International Conference on Affective Computing and Intelligent Interaction (ACII)}, pages 65--70. IEEE.

\bibitem[Dai et~al., 2019]{dai2019second}
Dai, T., Cai, J., Zhang, Y., Xia, S.-T., and Zhang, L. (2019).
\newblock Second-order attention network for single image super-resolution.
\newblock In {\em Proceedings of the IEEE/CVF conference on computer vision and pattern recognition}, pages 11065--11074.

\bibitem[He et~al., 2016]{he2016deep}
He, K., Zhang, X., Ren, S., and Sun, J. (2016).
\newblock Deep residual learning for image recognition.
\newblock In {\em Proceedings of the IEEE conference on computer vision and pattern recognition}, pages 770--778.

\bibitem[Hilles and Naser, 2017]{hilles2017knowledge}
Hilles, M.~M. and Naser, S. S.~A. (2017).
\newblock Knowledge-based intelligent tutoring system for teaching mongo database.(2017).

\bibitem[Hu et~al., 2019]{hu2019meta}
Hu, X., Mu, H., Zhang, X., Wang, Z., Tan, T., and Sun, J. (2019).
\newblock Meta-sr: A magnification-arbitrary network for super-resolution.
\newblock In {\em Proceedings of the IEEE/CVF conference on computer vision and pattern recognition}, pages 1575--1584.

\bibitem[Jing et~al., 2020]{jing2020feature}
Jing, W., Tian, F., Zhang, J., Chao, K.-M., Hong, Z., and Liu, X. (2020).
\newblock Feature super-resolution based facial expression recognition for multi-scale low-resolution faces.
\newblock {\em arXiv preprint arXiv:2004.02234}.

\bibitem[Lai et~al., 2017]{lai2017deep}
Lai, W.-S., Huang, J.-B., Ahuja, N., and Yang, M.-H. (2017).
\newblock Deep laplacian pyramid networks for fast and accurate super-resolution.
\newblock In {\em Proceedings of the IEEE conference on computer vision and pattern recognition}, pages 624--632.

\bibitem[Li et~al., 2017]{li2017reliable}
Li, S., Deng, W., and Du, J. (2017).
\newblock Reliable crowdsourcing and deep locality-preserving learning for expression recognition in the wild.
\newblock In {\em Proceedings of the IEEE conference on computer vision and pattern recognition}, pages 2852--2861.

\bibitem[Lim et~al., 2017]{lim2017enhanced}
Lim, B., Son, S., Kim, H., Nah, S., and Mu~Lee, K. (2017).
\newblock Enhanced deep residual networks for single image super-resolution.
\newblock In {\em Proceedings of the IEEE conference on computer vision and pattern recognition workshops}, pages 136--144.

\bibitem[Liu et~al., 2020]{liu2020facial}
Liu, Z., Li, L., Wu, Y., and Zhang, C. (2020).
\newblock Facial expression restoration based on improved graph convolutional networks.
\newblock In {\em MultiMedia Modeling: 26th International Conference, MMM 2020, Daejeon, South Korea, January 5--8, 2020, Proceedings, Part II 26}, pages 527--539. Springer.

\bibitem[Lukas et~al., 2016]{lukas2016student}
Lukas, S., Mitra, A.~R., Desanti, R.~I., and Krisnadi, D. (2016).
\newblock Student attendance system in classroom using face recognition technique.
\newblock In {\em 2016 International Conference on Information and Communication Technology Convergence (ICTC)}, pages 1032--1035. IEEE.

\bibitem[Nan et~al., 2022]{nan2022feature}
Nan, F., Jing, W., Tian, F., Zhang, J., Chao, K.-M., Hong, Z., and Zheng, Q. (2022).
\newblock Feature super-resolution based facial expression recognition for multi-scale low-resolution images.
\newblock {\em Knowledge-Based Systems}, 236:107678.

\bibitem[Ou and Wu, 2020]{ou2020efficient}
Ou, J. and Wu, H. (2020).
\newblock Efficient human pose estimation with depthwise separable convolution and person centroid guided joint grouping.
\newblock In {\em Pattern Recognition and Computer Vision: Third Chinese Conference, PRCV 2020, Nanjing, China, October 16--18, 2020, Proceedings, Part II}, pages 626--638. Springer.

\bibitem[Shao and Cheng, 2021]{shao2021fcnn}
Shao, J. and Cheng, Q. (2021).
\newblock E-fcnn for tiny facial expression recognition.
\newblock {\em Applied Intelligence}, 51:549--559.

\bibitem[Szegedy et~al., 2015]{szegedy2015going}
Szegedy, C., Liu, W., Jia, Y., Sermanet, P., Reed, S., Anguelov, D., Erhan, D., Vanhoucke, V., and Rabinovich, A. (2015).
\newblock Going deeper with convolutions.
\newblock In {\em Proceedings of the IEEE conference on computer vision and pattern recognition}, pages 1--9.

\bibitem[Tang et~al., 2019]{tang2019design}
Tang, J., Zhou, X., and Zheng, J. (2019).
\newblock Design of intelligent classroom facial recognition based on deep learning.
\newblock In {\em Journal of Physics: Conference Series}, volume 1168, page 022043. IOP Publishing.

\bibitem[T{\"u}men et~al., 2017]{tumen2017facial}
T{\"u}men, V., S{\"o}ylemez, {\"O}.~F., and Ergen, B. (2017).
\newblock Facial emotion recognition on a dataset using convolutional neural network.
\newblock In {\em 2017 International Artificial Intelligence and Data Processing Symposium (IDAP)}, pages 1--5. IEEE.

\bibitem[Wu et~al., 2021]{wu2021practical}
Wu, G., Jiang, J., Liu, X., and Ma, J. (2021).
\newblock A practical contrastive learning framework for single image super-resolution.
\newblock {\em arXiv preprint arXiv:2111.13924}.

\bibitem[Zhang et~al., 2018a]{zhang2018image}
Zhang, Y., Li, K., Li, K., Wang, L., Zhong, B., and Fu, Y. (2018a).
\newblock Image super-resolution using very deep residual channel attention networks.
\newblock In {\em Proceedings of the European conference on computer vision (ECCV)}, pages 286--301.

\bibitem[Zhang et~al., 2019]{zhang2019bridging}
Zhang, Y., Liu, T., Long, M., and Jordan, M. (2019).
\newblock Bridging theory and algorithm for domain adaptation.
\newblock In {\em International conference on machine learning}, pages 7404--7413. PMLR.

\bibitem[Zhang et~al., 2018b]{zhang2018residual}
Zhang, Y., Tian, Y., Kong, Y., Zhong, B., and Fu, Y. (2018b).
\newblock Residual dense network for image super-resolution.
\newblock In {\em Proceedings of the IEEE conference on computer vision and pattern recognition}, pages 2472--2481.

\bibitem[Zhu et~al., 2021]{zhu2021dynamic}
Zhu, M., Han, K., Wu, E., Zhang, Q., Nie, Y., Lan, Z., and Wang, Y. (2021).
\newblock Dynamic resolution network.
\newblock {\em Advances in Neural Information Processing Systems}, 34:27319--27330.

\end{thebibliography}

\end{document}